\documentclass[conference]{IEEEtran}
\IEEEoverridecommandlockouts
\usepackage{cite}
\usepackage{amsmath,amssymb,amsfonts}
\usepackage{algorithmic}
\usepackage{graphicx}
\usepackage{textcomp}
\usepackage{xcolor}
\def\BibTeX{{\rm B\kern-.05em{\sc i\kern-.025em b}\kern-.08em
    T\kern-.1667em\lower.7ex\hbox{E}\kern-.125emX}}
\begin{document}

\title{Predicting Alzheimer’s Disease by Hierarchical Graph Convolution from\\Positron Emission Tomography Imaging
\thanks{* Joint first authors}
}

\author{\IEEEauthorblockN{Jiaming Guo*}
\IEEEauthorblockA{\textit{Yuanpei College} \\
\textit{Peking University}\\
Beijing, China \\
markguo1998@gmail.com}
\and
\IEEEauthorblockN{Wei Qiu*}
\IEEEauthorblockA{\textit{Yuanpei College} \\
\textit{Peking University}\\
Beijing, China \\
qiuweipku@gmail.com}
\and
\IEEEauthorblockN{Xiang Li*}
\IEEEauthorblockA{\textit{Department of Radiology} \\
\textit{Massachusetts General Hospital}\\
Boston, MA \\
xli60@mgh.harvard.edu}
\and
\IEEEauthorblockN{Xuandong Zhao}
\IEEEauthorblockA{\textit{College of Computer Science} \\
\textit{Zhejiang University}\\
Hangzhou, China \\
csxuandongzhao@gmail.com}
\and
\IEEEauthorblockN{Ning Guo}
\IEEEauthorblockA{\textit{Department of Radiology} \\
\textit{Massachusetts General Hospital}\\
Boston, MA \\
Guo.Ning@mgh.harvard.edu}
\and
\IEEEauthorblockN{Quanzheng Li}
\IEEEauthorblockA{\textit{Department of Radiology} \\
\textit{Massachusetts General Hospital}\\
Boston, MA \\
li.quanzheng@mgh.harvard.edu}
}

\maketitle

\begin{abstract}
Imaging-based early diagnosis of Alzheimer Disease (AD) has become an effective approach, especially by using nuclear medicine imaging techniques such as Positron Emission Topography (PET). In various literature it has been found that PET images can be better modeled as signals (e.g. uptake of florbetapir) defined on a network (non-Euclidean) structure which is governed by its underlying graph patterns of pathological progression and metabolic connectivity. In order to effectively apply deep learning framework for PET image analysis to overcome its limitation on Euclidean grid, we develop a solution for 3D PET image representation and analysis under a generalized, graph-based CNN architecture (PETNet), which analyzes PET signals defined on a group-wise inferred graph structure. Computations in PETNet are defined in non-Euclidean, graph (network) domain, as it performs feature extraction by convolution operations on spectral-filtered signals on the graph and pooling operations based on hierarchical graph clustering. Effectiveness of the PETNet is evaluated on the Alzheimer's Disease Neuroimaging Initiative (ADNI) dataset, which shows improved performance over both deep learning and other machine learning-based methods.
\end{abstract}

\begin{IEEEkeywords}
Positron Emission Topography, Graph Convolution Network
\end{IEEEkeywords}

\section{Introduction}
\begin{figure*}
\centering
\includegraphics[width =\textwidth]{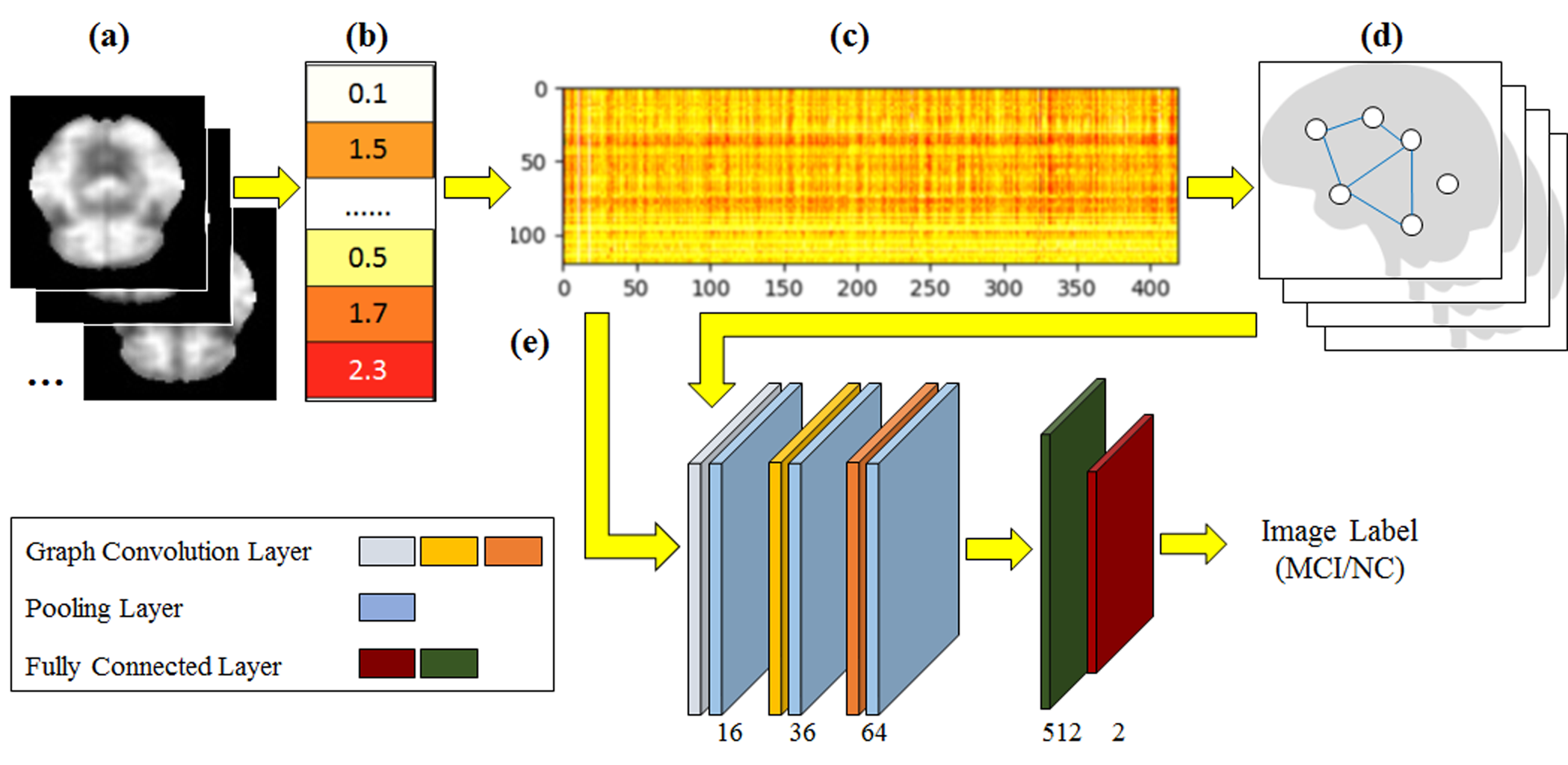}
\caption{Algorithmic pipeline of PETNet. (a) input PET images; (b) extracted average signals on ROIs; (c) aggregation of extracted signals across all subjects (x-axis) in each ROI (y-axis); (d) graph inferred from group-wise signals and its hierarchical clustering; (e) structure of graph convolution network, where both the extracted signals and the inferred graph are used as input.}
\label{fig:Fig1}
\end{figure*}

Imaging studies on human brain reveal that organization architecture of brain network forms a consistent pattern which is related to its cognition, behavior, and diseases \cite{b1}. In the field of analysis and diagnosis of Alzheimer's Disease (AD), studies from various imaging techniques including Positron Emission Tomography (PET) \cite{b2}, Magnetic Resonance Imaging (MRI) \cite{b3}, functional imaging \cite{b4}, diffusion imaging \cite{b5}, or a combination of these \cite{b6} have all confirmed that AD is associated with dysfunction of brain connectivity, thus considered as a “connectopathy” (connectivity disorder) \cite{b7}. Brain network analysis based on graph theory and topology has achieved good performance in diagnosis and staging of AD \cite{b8}, and group-wise network atlas have been inferred from multiple imaging modalities \cite{b9,b10}.\\
From a computation and imaging analytics perspective, both machine learning methods \cite{b11,b12} have achieved tremendous success in the characterization and classification of AD. The intrinsic nature of connectivity disorder of AD further motivates studies that can identify the optimized representation for brain images \cite{b13}, which can be generalized as the task of learning low-dimensional manifold embedded in a high-dimensional space \cite{b14}. While deep learning methods such as Convolutional Neural Network (CNN) can effectively learn the lower-to-higher representation of images and be used for AD classification \cite{b15,b16}, it is important to incorporate irregular structures (comparing with regular image grid) of graph into the deep learning scheme \cite{b17}. By defining PET data extracted from Regions of Interest (ROIs) as signals on the nodes of a graph, we can perform signal filtering and representation learning on graph, similar to the signal filtering and feature extraction (e.g. through convolution filters) in Euclidean space. Inspired by the algorithmic architecture proposed by Defferrard et. al \cite{b18} which learns localized spectral filters from the given graph and performs graph filtering through Chebyshev polynomial approximation, in this work we develop and implement a PET image analytics framework, the PETNet, to learn graph-based features and an effective classification system which can perform early diagnosis of Alzheimer’s disease i.e. prediction and staging of Mild Cognitive Impairment (MCI). 

\section{Material and Method}
As visualized in Fig.1 for the algorithmic pipeline, PETNet consists of four steps including 3D PET image ~\ref{fig:Fig1} (a) to ROI-based signal conversion (b), graph inference and hierarchical clustering (d), and a graph convolutional network for signal-to-label prediction (e). 

\subsection{Data acquisition, preprocessing and signal extraction}
We use the Florbetapir (F18-AV-45) PET images provided by Alzheimer's Disease Neuroimaging Initiative 2 (ADNI2) dataset (dynamic 3D scan, four 5-min frames, 50-70 min post injection) \cite{b19} for model validation and evaluation. As in this study our goal is to perform early AD diagnosis from PET images, we use patients who has been diagnosed as “Early Mild Cognitive Impairment” (EMCI, 131 subjects) or LMCI (Late Mild Cognitive Impairment, 96 subjects), as well as normal controls (NC, 100 subjects) for model input. For binary classification task, we combine EMCI and LMCI as “MCI positive” versus normal control (i.e. MCI negative) to differentiate disease and normal population. For multiclass classification task, we use the three labels simultaneously. Voxel values on PET images are defined by standard-uptake-value ratios (SUVR).
PET images are co-registered, averaged across frames, and standardized according to preprocessing protocol of ADNI-2 \cite{b19}. PET images are then registered to the standard 2mm-MNI152 space based on their corresponding MR images, with the shape of (91, 109, 91). Using AAL-2 atlas \cite{b20} consisting 120 Regions of Interest (ROI), we extract ROI-averaged PET signals from each individual 3D PET image, and aggregate extracted signals from all subjects into a data matrix of shape (120, 327). It should be noted that selection of atlas for ROI definition is fully customizable in PETNet, where the possibility of using other atlases are discussed later in the result section.
\subsection{Graph inference, hierarchical clustering and graph convolution network}
In the proposed PETNet framework, we are interested in a supervised learning task where input signal $x$, defined on a constant graph structure G consisting of n nodes, are used to predict its corresponding output $y$. As constant $G$ needs to be defined prior to the analysis, we infer the group-wise graph from signal matrix by its correlation. In other words, group-wise patterns of amyloid deposition among different regions is used to assist the model in performing signal filtering, as it reflects underlying short/long range metabolic connectivity and possible pathological progression pathways \cite{b21}. In the current framework a single graph G inferred from both patient and normal controls is used to characterize the common connectivity patterns, and we apply a threshold $(0.7)$ on the correlation values on G to make it sparse. Other graph inference methods can also be used based on different premise of data distribution. With a given $G$, the graph convolution filtering can be defined in spectral domain \cite{b22}: 
\begin{equation}\label {Eq:1}
\begin{aligned}
&G_L(x: \theta)=\sum_{k=0}^{K-1}{\theta_k U \Lambda^k U^T x} \\
\end{aligned}
\end{equation}
where graph Laplacian $L$ of $G$ can be diagonalized by its Fourier basis $U \in \mathbb{R}^{n \times n}$ such that $L=U \Lambda U^T$ and $\Lambda=diag({\lambda_1,\lambda_1,...,\lambda_n})$. The first $K$ number out of the totally $n$ basis, termed “kernel”, is used for filtering. According to \cite{b22}, kernel size $K$ exactly corresponds to the maximum shortest distance (“hops”) between two nodes on the graph, thus Eq.\ref{Eq:1} is localized to the $K-th$ neighborhood of each node. In other words, we can make analogous of kernel size $K$ to the field of view (FOV) of convolution kernels (e.g. $3 \times 3$ or $5 \times 5$) in classic CNNs. $\theta$ is the learnable coefficients (i.e. convolutional kernel parameters). Notice that since $G$ is constant through computation, its Laplacian and corresponding Fourier basis needs only be calculated once. Yet the matrix computation operation in Eq.\ref{Eq:1} still involves $O(n^2)$ complexity, thus we further approximate it by Chebyshev polynomial in same way as introduced in \cite{b18}: 
\begin{equation}\label {Eq:2}
\begin{aligned}
&G_L(x: \theta)=\sum_{k=0}^{K-1}{\theta_k T_k(L) x} \\
\end{aligned}
\end{equation}
where $T_k$ is the $k-th$ Chebyshev polynomial (to the total number of $K$) of $L$, and $\theta$ becomes coefficient for each polynomial. As Chebyshev approximation can be recursively computed, computation complexity reduces to $O(K \times |E|)$, $|E|$ is the total number of edges. $|E|$ is normally much smaller than $n^2$, especially for sparse graphs.
\subsection{Image-based method (CNN) for performance comparison}
CNN has been commonly applied for medical image analysis, including prediction of AD from PET imaging \cite{b16}. Due to the limitation in sample size of medical images, transfer learning techniques (i.e. pre-training) are usually needed during the training phase, especially for networks with complex structures. As pre-training is based on existing large image databases composed of 2D natural images (such as ImageNet), the network used for analyzing 3D medical images has to be limited to 2D input. In this work we use the same heuristics as in \cite{b16}, where a ResNet-50 \cite{b26} is trained by using 16 evenly-spaced z-sections from the 3D volumetric image, distributed into a 4×4 grid. The network is either trained from raw (initialized by random weights) or pre-trained (initialized by training on ImageNet).
\section{Result}
\subsection{Performance comparison}
\begin{figure}
\centering
\includegraphics[width =\columnwidth]{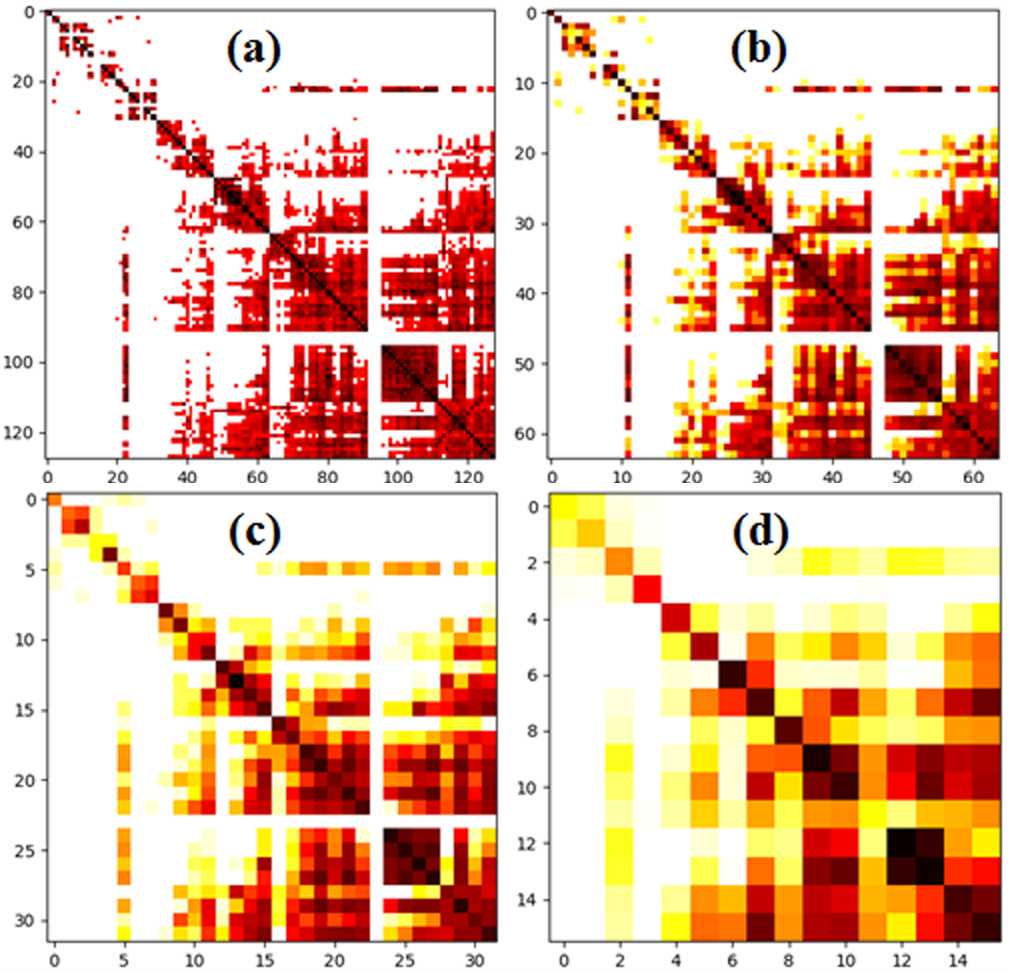}
\caption{Adjacency matrices of the (a) correlation-inferred graph among 120 regions; (b-d) first-to-third level clustering results, consisting of 60, 30 and 15 nodes respectively.}
\label{fig:Fig2}
\end{figure}

Performance evaluation of PETNet and comparison methods is based on 5-folds cross-validation ($80\%$ for training the model, $20\%$ for testing), repeated for 10 times. Graph structure for PETNet is inferred from the training data in each run and clustered by hierarchical clustering method, as visualized by their adjacency matrices in ~\ref{fig:Fig2}.
Commonly-used machine learning methods including XGBoost \cite{b27} and SVM with RBF kernel are also tested using the same ROI-based signal matrix as input. In order to investigate how the graph structure contribute to classification, we test PETNet with its graph replaced by:
\begin{enumerate}
\item An empty graph with only self-loops, which makes its Laplacian matrix $L$ equals to zero. In such case its graph convolution layers degrade to a fully connected layers, as spectral of an empty graph degrades to a single eigenvector.
\item Random graphs constructed by setting a fixed number of randomly-selected edges to random non-zero weights. We keep the number of edges in random graphs the same with graph inferred from correlation for comparability.
\end{enumerate}

\begin{figure}
\centering
\includegraphics[width =\columnwidth]{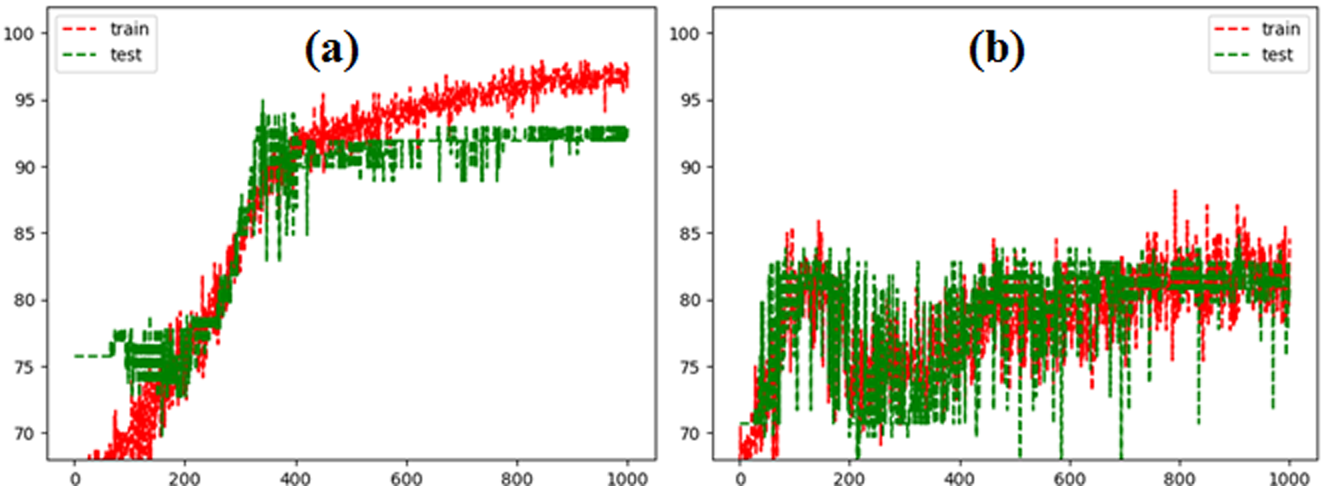}
\caption{Learning curve of (a) PETNet using correlation-inferred graph, and (b) PETNet using random graph, for one experiment run of ADNI binary classification task.}
\label{fig:Fig3}
\end{figure}

From the performance comparison in Table 1 “2-Classes” (i.e. MCI/NC) column, it can be observed that PETNet can achieve same-level performance with state-of-art pre-trained ResNet-50 method, where similar prediction accuracy for other dataset using ResNet-50 has been reported in \cite{b16}. It should be noted that without pre-training, performance of image-based deep learning decreased significantly (from $95\%$ down to $83\%$, p<0.01 for 10 experiments) because of the lack of samples to fully train such a complex model (50 layers). In contrast, PETNet uses much less number of mode parameters (5 layers in total), does not need pre-train, and uses much shorter time for training. Further, PETNet using empty/random graph results in lowered performance as well as unstable training process, as also illustrated by the learning curve in ~\ref{fig:Fig3}.\\

In addition, as patients diagnosed with MCI are staged by early stage (EMCI) and late stage (LMCI), we are interested in whether the two sub-groups can be differentiated. Using the same experiment setting, we evaluate these methods on the 3-labels prediction task. Both XGBoost and SVM are extended to their multi-class versions. Results are summarized in Table 1, column "3-classes" (i.e. EMCI/LMCI/NC). It can be found that while both ResNet and PETNet can achieve good performance for binary classification, the task of MCI staging is more difficult. ResNet mislabeled most of the EMCI patients to either NC or LMCI, which reduces its classification accuracy from $95\%$ to $65\%$. PETNet shows higher prediction power over other method and avoids the problem of EMCI mis-classification. Further, both ResNet without pre-training and PETNet defined on empty graph suffer from over-fitting, because of the lack of training samples, and the fact that PETNet on empty graph is effectively composed of all fully-connected layers without dropouts in its first three layers.

\begin{table}[]
\caption {Performance comparison among the proposed PETNet (and its variations), ResNet50 with/without pre-training, as well as other machine learning methods. Methods with the best performance for the two tasks (2-classes for MCI/NC classification, and 4-classes for EMCI/LMCI/NC classification) are highlighted in bold.} 
\centering
\begin{tabular}{ccc}
Method & \begin{tabular}[c]{@{}c@{}}Accuracy\\ (2-Classes)\end{tabular} & \begin{tabular}[c]{@{}c@{}}Accuracy\\ (3-Classes)\end{tabular} \\
PETNet & 93\% &{\textbf{77\%}} \\
PETNet (empty graph) & 88\% & 55\% \\
PETNet (random graph) & 86\% & 64\% \\
ResNet (without pre-training) & 83\% & 58\% \\
ResNet (with pre-training) &{\textbf{95\%}} & 65\% \\
XGBoost & 88\% & 62\% \\
SVM & 69\% & 57\%                                                          
\end{tabular}
\end{table}

\subsection{Discussion of atlas selection and kernel size parameter on model performance}
An important premise of PETNet is that we can effectively represent 3D volumetric PET images by its region-wise definition (i.e. averaged signals in ROIs). While ROI-based analysis has been commonly used for brain imaging studies, the selection of atlas and its corresponding ROI definitions can potentially affect model performance. Thus, in addition to the AAL-2 atlas (120 regions), we also tried Harvard-Oxford atlas (69 regions), MMP atlas (180 regions) and Power atlas (264 seed points with a pre-defined radius) for the analysis of ADNI dataset. Following the same experiment settings, it is found that performance of the proposed framework is very stable (accuracy changes $<1\%$) using different atlas. \\
PETNet features an important model parameter of kernel size K, which indicates the premise on effective neighborhood distance (i.e. K-hop) of the graph. A larger K will incorporate nodes that are more far apart into filtering operation, at the cost of increasing model complexity and training time. The default value of $K (25)$ is used in this work based on both parameter tuning and experience, as for the 120-nodes graph, a large portion of the nodes can be connected within a shortest distance of 25. Experiments on ADNI dataset using different values of K show a near-monotonic trend between K and model performance, where classification accuracy of PETNet is $84\%$ for $K=5$, $86\%$ for $K=10$ and $95\%$ for $K=40$.

\section{Conclusion and Discussion}
Preliminary results from PETNet show that deep learning based on graph-based representation can offer a more flexible and computational inexpensive approach for medical image analysis comparing with voxel-level modeling. An important premise of this work is that imaging data can be better modeled on a learnable graph, based on biological and anatomical evidence that brain regions physically/geometrically distant apart can be consistently correlated for their structural and functional properties including amyloid burden, cortical thickness, and cognitive roles. Such properties are less evidential in natural images, where distribution of pixel values are locally governed. The critical question for further development of graph-based analysis including the current PETNet framework is thus the graph construction/inference, which is an important, well-discussed yet still inconclusive topic in neuroimaging and medical image analysis in general. A specific lucrative approach is to incorporate graph inference process into the classification framework in order to learn the optimized graph definition. Other approaches including metrics learning (e.g. weight among different graph inference methods) and manifold learning can also be applicable.

\end{document}